\documentclass[conference]{IEEEtran}

\usepackage{graphicx}
\usepackage{amsmath,amssymb}
\usepackage{booktabs}
\usepackage{multirow}
\usepackage{cite}
\usepackage{url}
\usepackage{xcolor}
\usepackage[hidelinks]{hyperref}
\usepackage[caption=false,font=footnotesize]{subfig}
\usepackage{xurl}

\newcommand{\method}{StreamDAM}

\newcommand{\etal}{et al.\ }
\newcommand{\JF}{$\mathcal{J}\&\mathcal{F}$}
\newcommand{\best}[1]{\textbf{#1}}
\newcommand{\refr}[1]{\textcolor{gray}{#1}}

\title{StreamDAM: Presence-Aware Memory for Real-Time Streaming
Video Object Segmentation}

\author{
\IEEEauthorblockN{Xiang Chen}
\IEEEauthorblockA{
Global College\\
Shanghai Jiaotong University\\
Shanghai, China\\
xiangch@umich.edu
}
}

\begin{document}
\maketitle

\begin{abstract}
Quality-tier video object segmentation (VOS) trackers such as DAM4SAM top
accuracy leaderboards, but they are measured offline, one frame at a time with no
clock. Under an honest streaming protocol at 30 frames per second, where a frame
that misses its budget is served the last mask already computed, the winner
collapses: the rich memory that makes it accurate is too slow to keep up, and
what it emits is blind to whether the object is even present. We trace both
failures to one place, the tracker's memory pipeline, and rebuild it for
streaming. \method{} makes the memory machinery itself run at frame rate through
in-model optimization rather than a bolted-on fallback, and governs it with a
single learned presence signal that decides what enters memory, how far back the
tracker reads, when to withhold output, and when to re-detect. A mechanism
analysis shows why a fixed policy cannot win: the control that helps when an
object truly disappears is the one that hurts when it is merely hard to see, so
the choice must be made per frame. Across four benchmarks and five modern
baselines, \method{} is the strongest streaming tracker, recovers nearly all of
the offline model's accuracy under the clock, and on the hardest content exceeds
the offline model it is built from.
\end{abstract}


\section{Introduction}
\label{sec:introduction}

Promptable video object segmentation (VOS) has been transformed by memory-based
foundation trackers: SAM2~\cite{ravi2024sam2} and its distractor-aware refinement
DAM4SAM~\cite{videnovic2025dam4sam} top accuracy leaderboards by carrying a rich
memory of past frames. These systems are, however, almost always measured
\emph{offline}: each frame is processed to completion before the next is read, as
if the clock stopped. Real deployment in robotics, augmented reality, and video
editing has a frame clock. At 30 frames per second a new frame arrives roughly
every 33 milliseconds whether or not the tracker has finished the previous one.

This gap is sharp, not academic. Once a tracker's per-frame cost crosses the
budget it can no longer serve every frame, and the transition is abrupt: a
tracker only a hair over the budget can be forced to hold a stale mask on almost
every frame of a fast-changing sequence, and its streamed score
plummets~\cite{yang2024samurai}. DAM4SAM-L runs well over the budget and misses
everywhere; streamed under the zero-order-hold rule of the VOT real-time
protocol~\cite{kristan2017vot,kristan2023vots}, its offline lead over cheaper
models evaporates and it falls below a distilled efficiency tracker.

Two philosophies respond. \emph{Efficiency-first} trackers such as
EdgeTAM~\cite{edgetam} and EfficientTAM~\cite{efficienttam} distill SAM2 to fit
the budget, trading peak quality for speed; \emph{quality-first} trackers keep the
heavy memory and miss the clock. We argue for a third path. The two streaming
failures of a quality-first model are not separate problems bolted together: the
same memory pipeline is at once too slow for the frame clock and blind to whether
the object is present. \method{} rebuilds that one pipeline so its machinery runs
at frame rate and is governed, per frame, by a single presence signal, making the
same checkpoint both real-time and more accurate on the hard, distractor- and
disappearance-heavy content where its memory has structural advantage.

\noindent\textbf{Contributions.}
\begin{itemize}\itemsep0pt\parskip0pt
\item We locate the streaming failure of quality-tier VOS at a single place, the
tracker's memory pipeline, which is at once too slow for the frame clock and blind
to whether the object is present.
\item We make that memory machinery run at frame rate inside the model, with
emitted masks identical to the offline tracker: a graph-captured backbone,
host-synchronization removal, and resolution-capped distractor introspection, so
no accuracy is traded for speed (Sec.~\ref{sec:speed}).
\item We show by mechanism analysis that one coupled memory control governs the
tracker's hardest behavior and that no fixed setting of it can win, then govern
it with a single learned presence signal that drives memory admission, the
recency window, output suppression, and re-detection
(Sec.~\ref{sec:mech}, Sec.~\ref{sec:head}).
\item We evaluate under a frozen streaming protocol against five modern baselines
on four benchmarks, with explicit offline and stale-reuse bounds and an honest
ledger of what did not work
(Sec.~\ref{sec:experiments}, App.~\ref{app:ledger}).
\end{itemize}

\section{Related Work}
\label{sec:related_work}

\noindent\textbf{Memory-based VOS.} Space-time memory
networks~\cite{oh2019stm} and their efficient successors
XMem~\cite{cheng2022xmem} and Cutie~\cite{cheng2024cutie} made read/write memory
the dominant VOS paradigm. SAM2~\cite{ravi2024sam2} added a streaming memory bank
to promptable segmentation, and DAM4SAM~\cite{videnovic2025dam4sam} augments it
with a distractor-resolving memory (DRM) that inserts hard frames as conditioning
tokens and wins VOTS-class benchmarks. We take DAM4SAM as our base and dissect
the memory machinery that streaming stresses.

\noindent\textbf{Efficient VOS.} EdgeTAM~\cite{edgetam} and
EfficientTAM~\cite{efficienttam} distill or restructure SAM2's encoder for
on-device speed. They meet the clock but shed the heavy memory that helps on hard
content, and they are our strongest streaming baselines.

\noindent\textbf{Training-free SAM2 modifications.}
SAM2Long~\cite{sam2long} adds uncertainty-branched memory pathways and
SAMURAI~\cite{yang2024samurai} Kalman-filtered mask selection, both training-free.
We use SAMURAI-style selection as an ablation, where it is a measured wash on
DAM4SAM, and, unlike these, add one small \emph{learned} component only where the
training-free signal is provably insufficient.

\noindent\textbf{Streaming perception and real-time protocols.} Li~\etal
\cite{li2020streaming} defined \emph{streaming accuracy}, coupling latency and
quality under a clock; the VOT and VOTS real-time
protocol~\cite{kristan2017vot,kristan2023vots} scores trackers under a deadline
with stale-output reuse; classic video work amortizes compute across frames, as
in Deep Feature Flow~\cite{zhu2017dff} and Clockwork
Convnets~\cite{shelhamer2016clockwork}. We adopt the streaming and stale-reuse
protocol and target the model itself, not inter-frame amortization.

\section{StreamDAM}
\label{sec:method}

Both streaming failures live in one machine, the memory pipeline: it is too slow
for the frame clock and blind to whether the object is present. \method{} rebuilds
it for streaming. We formalize the protocol (Sec.~\ref{sec:protocol}) and the
memory read (Sec.~\ref{sec:mech}), stream the pipeline (Sec.~\ref{sec:speed}),
learn a causal presence estimate (Sec.~\ref{sec:presence}), and use it to govern
the memory (Sec.~\ref{sec:head}).

\subsection{Problem setting and streaming protocol}
\label{sec:protocol}
Frame $t$ is released at time $t\Delta$ under a fixed clock of period
$\Delta = 1/f$, the per-frame budget, and a frame started at $\sigma$ finishes its
mask at $\sigma + c$ for compute cost $c$. At each deadline the tracker emits the
newest finished output, serving the last completed mask on a miss, a zero-order
hold, exactly the VOT real-time rule~\cite{kristan2017vot,kristan2023vots}:
\begin{equation}
\hat m_t = m_{\pi(t)}, \quad
\pi(t) = \max\{\, k : \text{frame } k \text{ ready by } t\Delta \,\}.
\label{eq:emit}
\end{equation}
With the compute-to-budget ratio $\rho = c/\Delta$ over steady-state compute $c$,
the served mask stays fresh while $\rho \le 1$, but once $\rho$ exceeds one the
tracker falls behind and $\pi(t) < t$ on a growing share of frames. The dependence
is a cliff: a tracker a hair above $\rho = 1$ holds a stale mask on nearly every
frame of fast content.

\subsection{The memory pipeline}
\label{sec:mech}
\begin{figure}[t]
\centering
\includegraphics[width=0.92\columnwidth]{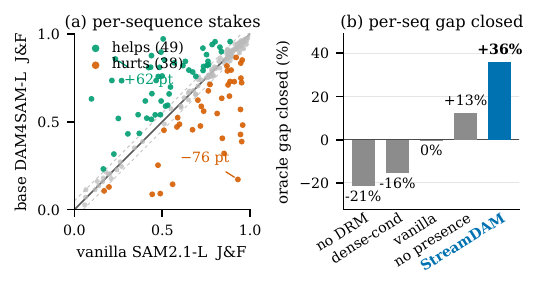}
\caption{Per-sequence stakes and payoff of DAM4SAM's fixed memory on held-out
MOSE-train, by J\&F. \textbf{(a)} Each dot is one sequence, base DAM4SAM-L against
vanilla SAM2.1-L: the memory helps some and hurts others by huge margins, swinging
by more than sixty points either way and tens on average. \textbf{(b)} Against a
per-sequence oracle of the better design, fixed policies close at most a fifth of
the gap and several widen it, while \method{} closes over a third, several times
the best fixed policy. Presence must be decided per frame.}
\label{fig:mech}
\end{figure}

DAM4SAM reads a memory bank $B_t$ whose entries $j$ store encoded features $e_j$
and the decoded mask $m_j$. Each frame it encodes $x_t$ to $f_t = \phi(x_t)$
through the backbone $\phi$, assembles a conditioning set $M_t$, attends $f_t$
against it to decode $m_t$, and writes back a new entry:
\begin{equation}
M_t = C \,\cup\, \{\, e_j \in B_t : |m_j| > 0 \,\wedge\, j \in W_r(t) \,\},
\label{eq:cond}
\end{equation}
with prompt anchors $C$. The presence filter $|m_j| > 0$ drops entries with an
empty stored mask, and the temporal stride $r$ thins the recent window
$W_r(t) = \{t-1, t-1-r, \dots\}$; together they form one coupled recency control.
This pipeline is both bottlenecks: the backbone $\phi$ dominates compute and is
flat in video length, so it sets $\rho$, while on held-out sequences the coupled
control alone carries DAM4SAM's split behavior, helping some sequences and hurting
others though its distractor additions carry none of it. It is right on a genuine
disappearance and wrong on present-but-low-confidence frames, no fixed setting
winning both (Fig.~\ref{fig:mech}): presence must be known per frame.

\subsection{Streaming the pipeline}
\label{sec:speed}
The latency is all memory machinery, reclaimed by three in-model changes that are
bit-identical by construction (Table~\ref{tab:latency}): graph-capturing the Hiera
backbone with \texttt{torch.compile} and CUDA-graph capture; removing host
synchronization through a GPU-resident mask, byte-packed transfers, and on-device
pixel counts; and capping the distractor introspection, whose connected-components
cost scales as $\min(R, s)^2$ in mask side-length $R$ under a cap $s$ while its
scale-invariant add decision is unchanged. These bring streamed compute below
budget, so $\rho \le 1$ save one oversize sequence, the honest residual flagged in
Sec.~\ref{sec:limits}. Speed leaves the accuracy bottleneck untouched.

\begin{table}[t]
\centering\small
\caption{Per-component latency of DAM4SAM-L and the effect of each in-model
optimization, in ms per frame at 1080p unless noted. All optimizations are
bit-identical by construction. Full profiles in App.~\ref{app:latency}.}
\label{tab:latency}
\setlength{\tabcolsep}{3.5pt}
\resizebox{\columnwidth}{!}{%
\begin{tabular}{lccc}
\toprule
Component / stage & pub.\ & optimization & after \\
\midrule
Hiera-L trunk        & 23.2 & CUDA-graph      & ${\sim}16.5$ \\
CPU host path        & 6.3  & copy elimination        & ${\sim}4.4$ \\
DRM introsp., 1080p tail & 6.1 & downsampling & ${\sim}0$ \\
mem-attn/dec/enc     & 8.6  & unchanged           & 8.6 \\
\midrule
\textbf{per-frame total} & \textbf{${\sim}38$} & & \best{$<32$ p95} \\
\bottomrule
\end{tabular}}
\end{table}

\subsection{Estimating presence}
\label{sec:presence}
The built-in presence test thresholds the object-score logit and fails in the
low-confidence band that matters most. We add a small causal estimator: a gated
recurrent unit (GRU) $g_\theta$ reads per-frame runtime scalars $u_t$ the tracker
already computes---logits, mask-overlap statistics, area ratios, absence
run-length---and carries a state $h_t$ to a calibrated presence probability
\begin{equation}
h_t = g_\theta(u_t, h_{t-1}), \qquad p_t = \sigma(w^\top h_t + b),
\label{eq:head}
\end{equation}
trained on disjoint held-out sequences against ground-truth presence
$y_t \in \{0,1\}$ with a class-balanced binary cross-entropy
\begin{equation}
\mathcal{L} = -\sum_t \big[\, \beta\, y_t \log p_t
+ (1-\beta)(1-y_t)\log(1-p_t) \,\big].
\label{eq:bce}
\end{equation}
Every operating threshold below is set on training data alone, and $g_\theta$
costs a negligible slice of the budget on CPU.

\subsection{Presence-governed memory control}
\label{sec:head}
The signal $p_t$ governs four decisions in \eqref{eq:cond}. \emph{Admission}
replaces the presence filter, a bank entry becoming eligible when its stored
estimate clears $\tau$:
\begin{equation}
j \in M_t \iff p_j > \tau .
\label{eq:admit}
\end{equation}
\emph{The recency window} is a binary stride mode $m^{\mathrm{w}}_t$ set by a
hysteresis machine on $p_t$, dense when it equals $1$ and held only for a burst
after a reappearance:
\begin{equation}
m^{\mathrm{w}}_t = 1 \iff p_t \ge p_{hi} \,\vee\,
\big( m^{\mathrm{w}}_{t-1} \wedge p_t \ge p_{lo} \wedge d_t < R \big),
\label{eq:window}
\end{equation}
with $d_t$ the frames since dense onset and the window sparse otherwise.
\emph{Output suppression} emits $\hat m_t = \varnothing$ when $p_t < \theta$, and
\emph{re-detection} relaxes \eqref{eq:admit} after $K$ suppressed frames to break
the self-reinforcing absent latch. Admission recovers the hurt sequences and the
burst protects the helped ones, together the first policy we found to close the
per-sequence gap no fixed setting reaches (Fig.~\ref{fig:mech}); the
other two are conservative safety governors
(App.~\ref{app:gates}, App.~\ref{app:repro}). One signal, four consumers, and no
fixed policy serves all four at once.

\section{Experiments}
\label{sec:experiments}

\noindent\textbf{Setup.} All runs use a single NVIDIA RTX 6000D GPU of the
Blackwell generation, with PyTorch 2.11 on CUDA 12.8 and bfloat16 inference, and
every latency we report is wall-clock on this GPU. The timed harness replays each
sequence under a 30\,fps clock: frame $t$ is released at $t/30$\,s, the tracker
always processes the newest released frame, and whenever a frame's budget passes
before a fresh mask is ready we emit the last computed mask, a zero-order hold
(ZOH), exactly the VOT real-time rule~\cite{kristan2017vot,kristan2023vots}. The
model and any CPU fast path run for real, so masks are faithful and the timeline
is driven by measured cost. We report each method both \emph{streamed}, its honest
score under the clock, and \emph{offline}, with oracle compute and no clock. Where
evaluation is parallelized we use several GPUs of the same model, but serving is
always single-GPU. All methods share this harness, are initialized uniformly from
the first-frame ground-truth mask as a fairness fix, and are scored with
single-object region-and-contour accuracy, \JF{}. We report \method{} as the mean
over three fresh-compile seeds, with seed variation below one point on three of
the four benchmarks and LVOS the exception.

\noindent\textbf{Datasets and baselines.} We evaluate on four benchmarks of
rising difficulty: DAVIS-2017 val~\cite{ponttuset2017davis}, a saturated
reference; LVOS val~\cite{hong2023lvos}, long videos; VOST
val~\cite{tokmakov2023vost}, heavy transformation and occlusion; and
\textbf{MOSE-hard}, sixty sequences curated from held-out
MOSE~\cite{ding2023mose} \emph{train} by ground-truth difficulty and frozen
before comparison, described in App.~\ref{app:moseh}. Baselines are recent:
EdgeTAM, EfficientTAM, Cutie~\cite{cheng2024cutie}, SAMURAI, and vanilla
SAM2.1-L. For the fast baselines streamed equals offline, verified with a zero
hold-fraction on every spot-check; SAMURAI and vanilla-L straddle or miss the
clock and degrade honestly.

\begin{figure}[t]
\centering
\includegraphics[width=0.70\columnwidth]{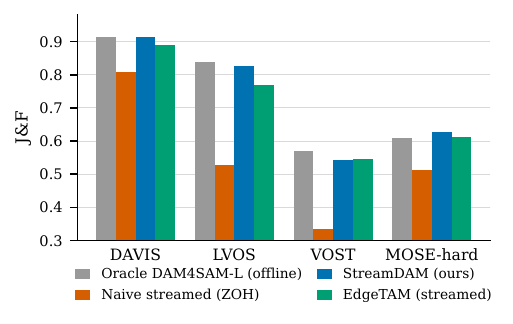}
\caption{The streaming cliff and its recovery. Naive stale-reuse streaming
collapses the offline model on every benchmark; \method{} recovers nearly all of
it, matches or leads the strongest streaming baseline EdgeTAM, and on MOSE-hard
surpasses the offline model.}
\label{fig:cliff}
\end{figure}

\noindent\textbf{Main result.} Table~\ref{tab:main} is the primary streamed
comparison, with explicit offline and stale-reuse bounds. \method{} is the
strongest streaming tracker on the mean, ahead of EdgeTAM and EfficientTAM. It
leads the two hardest panels, MOSE-hard and LVOS, ties VOST, and cedes only the
saturated DAVIS reference to the efficiency specialists, as expected. The
comparison against the offline bound is the headline: naive streaming sheds nearly
nineteen points of the offline model's mean accuracy, and \method{} recovers about
ninety-seven percent of that gap while meeting the same clock. On MOSE-hard its
\emph{streamed} score even surpasses the offline model, because the presence
governor improves the model itself, not merely how it is served
(Fig.~\ref{fig:cliff}).

\begin{table}[t]
\centering\small
\caption{Streamed 30\,fps \JF{}, single object, with offline and stale-reuse
bounds. Grey rows are references; bold marks the best streamed value per column.
\method{}, our method, is the mean over three fresh-compile seeds, seed spread
below one point except on LVOS.}
\label{tab:main}
\setlength{\tabcolsep}{3.2pt}
\resizebox{\columnwidth}{!}{%
\begin{tabular}{lccccc}
\toprule
Method, streamed 30\,fps & DAVIS & LVOS & VOST & MOSE-h & mean \\
\midrule
\refr{Oracle DAM4SAM-L, offline} & \refr{0.914} & \refr{0.838} & \refr{0.570} & \refr{0.610} & \refr{0.733} \\
\refr{\method{} offline, ref.}   & \refr{0.913} & \refr{0.832} & \refr{0.547} & \refr{0.627} & \refr{0.730} \\
\midrule
\best{\method{}} & \best{0.914} & \best{0.825} & 0.544 & \best{0.626} & \best{0.727} \\
EdgeTAM            & 0.891 & 0.769 & \best{0.545} & 0.612 & 0.704 \\
EfficientTAM       & \best{0.921} & 0.733 & 0.528 & 0.593 & 0.694 \\
vanilla SAM2.1-L   & 0.884 & 0.810 & 0.413 & 0.570 & 0.669 \\
Cutie              & 0.908 & 0.658 & 0.483 & 0.588 & 0.659 \\
SAMURAI            & 0.879 & 0.811 & 0.341 & 0.508 & 0.635 \\
\midrule
Naive DAM4SAM-L, ZOH & 0.809 & 0.527 & 0.335 & 0.513 & 0.546 \\
\bottomrule
\end{tabular}}
\end{table}

\noindent\textbf{Attribution.} An offline decomposition with full anchors
separates the two additions \method{} makes over the base tracker. The uniform
mask-initialization fix carries the DAVIS gain and part of MOSE-hard; the presence
governor carries the rest of MOSE-hard at a small cost on VOST, and does not buy
that gain by regressing elsewhere---a pre-registered cross-domain check finds no
family regressing beyond noise.

\noindent\textbf{Serving.} \method{} meets the clock, genuinely real-time rather
than a stale-reuse model in disguise: its stale-hold fraction is near zero apart
from one oversize sequence, and its compute stays within the frame budget at the
95th percentile (Table~\ref{tab:latency}).

\noindent\textbf{Negatives.} App.~\ref{app:ledger} records every rejected lever.
Input-resolution scaling is catastrophic on small objects: at reduced resolution
the bounding-box init grabs a distractor, a re-roll not a rescue. DRM bank hygiene
is refuted, since every variant trades away more on the help group than it
recovers on the hurt group. Compile-lottery variance is real and non-Gaussian;
pinning the autotune cache removes it for deployment and is the source of the LVOS
seed spread.

\section{Limitations and Conclusion}
\label{sec:limits}
\method{} has honest boundaries. LVOS carries compile-lottery variance, so the
pinned-cache remedy is a deployment step, not a proof; one oversize VOST sequence
still breaches the budget, an un-eliminated resolution tail; a large share of the
hurt group is moved by neither presence consumer; our protocol scores
single-object accuracy, leaving multi-object streaming open; and MOSE-hard is a
curated held-out subset of MOSE train, frozen before comparison, a fair but
non-standard panel we state plainly.

We conclude that the streaming cliff, not raw offline accuracy, is the right lens
for deploying quality-tier VOS, and that the productive response is to rebuild the
memory pipeline itself: make its machinery real-time inside the model and govern
it with a single presence signal. \method{} makes DAM4SAM-L real-time with
bit-identical outputs, localizes its hurt and help behavior to one coupled memory
control, and governs that control with a small learned presence head that,
uniquely among the policies we tried, recovers the hurt group without hurting the
help group---giving the strongest streaming tracker across four benchmarks and, on
the hardest, beating the offline model. Closing the small gap that remains
everywhere needs a learned regime selector over the memory composition, which
online causal signals cannot separate---the natural next step.

\bibliographystyle{IEEEtran}
\bibliography{references}

\clearpage
\appendices

\section{Negative-results ledger}
\label{app:ledger}
Every rejected lever, with evidence.
\begin{itemize}\itemsep1pt
\item \textbf{Input-resolution scaling (rejected).} L@896 on the 8 smallest-area
LVOS seqs (area frac $0.0003$--$0.0009$): mean \JF{} $0.470$ vs.\ $0.855$@1024
($-0.385$; $4/8$ catastrophic collapse); L@960 $-0.253$ with \emph{different}
seqs collapsing. Root cause: the bbox-prompt init grabs a distractor at reduced
resolution (frame-0 IoU $0.00$ vs.\ $0.88$@1024). Knife-edge, not gradual.
\item \textbf{DRM bank hygiene (refuted).} TTL expiry / cap / freeze-after all
trade help-group losses $\ge$ hurt-group gains; the DRM hurt/help split is a
per-sequence regime (only $1/15$ divergences within 10 frames of a DRM add;
add-time separability AUC $\le0.67$), so per-event add-veto is dead.
\item \textbf{Memory-bank shrinking (rejected).} LVOS $0.949\rightarrow0.434$.
\item \textbf{Aggressive/matched-fp admission $\tau$ (rejected).} The
emission-gate's matched-fp $\tau{=}0.46$ is net-exclusionary for admission (hurt
$-0.037$); the looser $\tau{=}0.34$ is correct.
\item \textbf{Head emission variant, embedding-similarity feature,
channels-last-on-compile, antialias-skip} (rejected: no gain / parity failure).
\item \textbf{Kalman-filter mask selection (wash).} Free, kept, claimed only as
an ablation row.
\end{itemize}

\section{MOSE-hard curation}
\label{app:moseh}
MOSE-hard is 60 sequences drawn from the held-out MOSE \emph{train} split
(disjoint from the presence-head train corpus), selected by objective GT-only
difficulty attributes---disappearance/reappearance events, number of
co-annotated distractors, object size, and video length---and frozen in a fixed
sequence list \emph{before} any method comparison. It concentrates exactly the
content where distractor-aware memory has structural advantage. It is a
non-standard panel and we report it as such.

\section{Per-component latency}
\label{app:latency}
Steady-state DAM4SAM-L (1080p): encoder $23.2$\,ms ($61\%$, flat in history), CPU
``everything-else'' $6.3$ ($\mathrm{p95}\ 10.4$; DRM alt-mask numpy + host syncs),
memory attention $4.6$, decoder $3.0$, memory encoder $1.0$. The trunk compile
saves $6.0$--$7.6$\,ms; CPU reclaim $1.6$--$2.2$\,ms; \texttt{hires\_cap}=512
removes the $6$--$8$\,ms 1080p DRM-introspection tail (scale-invariant decision,
bit-identical on $5/6$ seqs). Streamed compute p95 by panel: DAVIS $27.8$, LVOS
$29.1$, VOST $31.0$ (max $36.0$ on the $2560{\times}1920$ residual), MOSE-hard
$29.9$. A featurizer subtlety cost a campaign: the presence head's full-res
mask-centroid scan (not the embedding pool, which is $0.04$\,ms) was the hi-res
tail; exact marginal row/col sums make it $0.95$\,ms and bit-identical.

\section{Gate evidence and forensics}
\label{app:gates}
Absence gate $\theta{=}1.0$ improves MOSE-train accuracy by a little over one
point, is roughly neutral on DAVIS/LVOS-train, and costs under a point on VOST, an
accepted trade. Redetection nudge $K{=}5$ breaks the \texttt{is\_obj\_appearing}
self-reinforcing latch (memory encodes ``absent'' $\rightarrow$ 235-frame dead
runs); conservative guard, helps rarely, never hurts. Presence head operating
points (val): head@matched-fp false-absent $0.067$ vs.\ logit's $0.099$ at equal
hallucination; head@youden cuts false-present to $0.080$ vs.\ logit's $0.234$. The
residual failure class (\texttt{eeaaa67e}) moves for neither presence consumer.

\section{Reproducibility}
\label{app:repro}
Hardware: single NVIDIA RTX 6000D (Blackwell, sm\_120)
GPU, bf16 autocast, torch 2.11+cu128. Evaluation: timed simulator at 30\,fps, ZOH
on miss, 3 fresh-compile seeds; the LVOS seed spread is $0.035$ across fresh
compiles. A compile guard (\texttt{--assert\_warm\_ms 33}) aborts poisoned
launches and CPU thread pools are capped (numerics unchanged).

\end{document}